\documentclass[conference]{IEEEtran}
\IEEEoverridecommandlockouts

\usepackage{cite}
\usepackage{amsmath,amssymb,amsfonts}
\usepackage{algorithmic}
\usepackage{graphicx}
\usepackage{textcomp}
\usepackage{xcolor}
\usepackage{makecell}
\usepackage{amssymb}
\usepackage{pifont}
\usepackage{array} 
 \usepackage{multirow}

\def\BibTeX{{\rm B\kern-.05em{\sc i\kern-.025em b}\kern-.08em
    T\kern-.1667em\lower.7ex\hbox{E}\kern-.125emX}}
\begin{document}

\title{Meta-Imputation Balanced (MIB): An Ensemble Approach for Handling Missing Data in Biomedical Machine Learning}

\author{%
\IEEEauthorblockN{Fatemeh Azad}
\IEEEauthorblockA{\textit{University of Ljubljana}\\
\textit{Faculty of Computer and Information Science}\\
Ljubljana, Slovenia\\
fatemeh.azad@fri.uni-lj.si}
\and
\IEEEauthorblockN{Zoran Bosnić}
\IEEEauthorblockA{\textit{University of Ljubljana}\\
\textit{Faculty of Computer and Information Science}\\
Ljubljana, Slovenia\\
zoran.bosnic@fri.uni-lj.si}
\and
\IEEEauthorblockN{Matjaž Kukar}
\IEEEauthorblockA{\textit{University of Ljubljana}\\
\textit{Faculty of Computer and Information Science}\\
Ljubljana, Slovenia \\
matjaz.kukar@fri.uni-lj.si}
}

\maketitle

\begin{abstract}
Missing data represents a fundamental challenge in machine learning applications, often reducing model performance and reliability. This problem is particularly acute in fields like bioinformatics and clinical machine learning, where datasets are frequently incomplete due to the nature of both data generation and data collection. While numerous imputation methods exist, from simple statistical techniques to advanced deep learning models, no single method consistently performs well across diverse datasets and missingness mechanisms. This paper proposes a novel Meta-Imputation approach that learns to combine the outputs of multiple base imputers to predict missing values more accurately. By training the proposed method called Meta-Imputation Balanced (MIB) on synthetically masked data with known ground truth, the system learns to predict the most suitable imputed value based on the behavior of each method. 

We evaluate our method on tabular data under the Missing Completely at Random (MCAR) assumption using both direct metrics, where Mean Absolute Error (MAE) and Root Mean Squared Error (RMSE) are computed between imputed values and their corresponding original ground truth values in the artificially masked positions, and indirect metrics, which measure the RMSE of a target variable predicted by machine learning models trained on the imputed datasets. Across three benchmark datasets, the model achieved the lowest or near‑lowest RMSE and delivered stable downstream predictive performance, even when individual imputers varied in performance. Our work highlights the potential of ensemble learning in imputation and paves the way for more robust, modular, and interpretable preprocessing pipelines in real-world machine learning systems.
\end{abstract}

\begin{IEEEkeywords}
Missing data, imputation, machine learning, deep learning, meta-model.
\end{IEEEkeywords}

\section{Introduction}
Missing data is a common challenge in real-world applications across various fields, such as automated measuring (sensor data), healthcare, finance, and engineering. Missingness occurs when particular entries (feature values) in a dataset are unrecorded, incomplete, or corrupted.  In clinical diagnostics, for instance, a patient may have results for only a limited subset of available laboratory tests due to protocols, cost, or patient-specific factors, creating high-dimensional but sparsely populated datasets. Missing data can significantly impact machine learning models, resulting in biased estimates, reduced predictive power, and unreliable decisions --- a critical issue for sensitive applications like clinical decision support. For example, in medical studies, patients may avoid sensitive questions, leading to incomplete data. Missing data can significantly impact machine learning models, resulting in biased estimates, reduced predictive power, and unreliable decisions.
The three primary mechanisms of missingness, as defined by Little \& Rubin \cite{little2019statistical}, are:
	\begin{itemize}
		\item Missing Completely at Random (MCAR): The probability of a value being missing (missingness) is independent of both observed and unobserved variables. For example, if a lab instrument randomly fails and skips some measurements regardless of patient characteristics or values, the missingness is MCAR \cite{little2019statistical}.
		\item Missing at Random (MAR): The probability of missingness depends only on the observed variables, not the missing datum itself. For example, if all missing salary values occur for female employees, the missingness is considered MAR, as can be explained by the observed gender variable \cite{little2019statistical, zhou2024review}.
		\item Missing Not at Random (MNAR): The probability of missingness is related to itself or the value of some unobserved variables. For example, people with very high incomes may be less likely to report their wealth. Hence, there is no external observed factor explaining the missingness. If we do not know the income, we cannot explain why it is missing. This is the most challenging \noindent scenario as the missingness itself is informative \cite{zhou2024review, little2019statistical}.
	\end{itemize}
	
	Here, “observed variables” refers to features in the dataset that have recorded values, while “unobserved variables” are features or values that are missing or never collected. Addressing missing values is crucial for reliable and high-performing data analysis. Methodologies for handling missingness can be broadly classified into deletion and imputation. 

Imputation refers to the process of estimating and filling in missing values in a dataset, allowing for complete and uninterrupted data analysis and modeling. Traditional approaches for addressing missing data include deletion-based methods and statistical imputation techniques \cite{zhou2024review}, such as mean, median, or regression-based estimations. While these traditional methods are simple and computationally efficient, they fail to capture complex feature dependencies and often introduce biases. More advanced machine learning approaches, such as k-nearest neighbors (KNN) \cite{KNN}, decision trees \cite{decision_tree}, and support vector machines (SVM) \cite{svm_missing}, have been employed to enhance imputation quality. Deep learning techniques, including Variational Autoencoders (VAEs) \cite{vae_missing}, Generative Adversarial Networks (GANs) \cite{GAN}, and Recurrent Neural Networks (RNNs) \cite{rnn_missing}, have demonstrated significant improvements in imputing missing values, particularly in high-dimensional datasets. However, they often face scalability, interpretability, and efficiency challenges in real-world applications. Handling missing data becomes even more important and complex due to different missing data mechanisms \cite{allison2001missing}. Most imputation techniques fail to generalize across all these mechanisms.

This paper proposes a Meta-Imputation Balanced (MIB) framework for handling missing data using machine learning techniques to enhance accuracy and robustness. Unlike traditional methods that rely on a single imputation strategy, our framework combines multiple imputation techniques into a unified predictive model. The framework blends imputation strategies for each value by training a regression model on masked data and their imputed values. This approach offers a flexible, modular, and interpretable solution that was evaluated on benchmark datasets for imputation accuracy and predictive performance.

The rest of this paper follows this structure: In Section 2, we review existing literature on missing data imputation techniques, highlighting their strengths and limitations. In Section 3, we present our methodology in detail. In Section 4, we describe our evaluation setup. In Section 5, we present and analyze the results. Finally, in Section 6, we summarize our study and explore future research directions.

\section{Related Work}

Researchers have developed various methods to address missing data problems, from simple statistical techniques to complex deep-learning models. This section provides an overview that begins with traditional approaches and progresses to modern solutions.

\subsection{Traditional and ML Imputation Approaches}

Early techniques approached missing data with deletion or basic statistical imputations. These methods, such as listwise deletion \cite{zhou2024review} or mean \cite{article}, quickly remove missing entries or replace them with average values. However, they often distort data distributions and reduce predictive accuracy, especially when data is not missing completely at random (MCAR) \cite{allison2001missing}.

Machine learning methods address these limitations by modeling relationships between features. Murti et al. \cite{KNN} applied KNN techniques, imputing missing data based on nearby examples. Although this approach works well in low-dimensional settings, its performance and scalability significantly decrease as data dimensionality increases. Similarly, Feng et al. \cite{svm_missing} framed missing data imputation as a supervised learning problem and trained an SVM to fill in missing entries. Their method surpassed baseline imputation techniques such as mean and KNN imputations in preserving data structure and with higher accuracy. Matrix factorization \cite{matrix} decomposes the data matrix into latent factors, enabling the estimation of missing values by reconstructing the original matrix from these learned factors. 

Deep learning provides even more flexible and adaptable solutions for imputation tasks. Yoon et al. \cite{gain} suggested Generative Adversarial Imputation Networks (GAIN), which use adversarial training to train a generator and a discriminator to create realistic imputations and distinguish between observed and generated values. Their method included a hint mechanism, which helped the model to align generated values with feasible patterns in the data. Collier et al. \cite{vae_missing} extended Variational Autoencoders (VAEs) to address missing data problems by integrating binary masks that indicate missingness. This modification allowed their model to estimate distributions under both MCAR and MNAR conditions, producing more accurate imputations than standard VAEs. Becker et al. \cite{rnn_missing} proposed a Recurrent Neural Network (RNN) framework that predicts future time steps and then refines predictions using recent observations. Their architecture handled sequential data effectively, maintaining consistency in temporal tasks.
Other researchers modified network architectures directly. Śmieja et al. \cite{neural_net_missing} redefined how neurons process incomplete inputs by calculating expected activations over a distribution of possible values. This innovation enabled their models to learn directly from incomplete data without preprocessing. Przewieżlikowski et al. \cite{misconv} developed MisConv, which used mixtures of factor analyzers to model the uncertainty of missing pixels in Convolutional Neural Networks (CNNs). Their method preserved image structure and improved robustness in visual domains.

Muzellec et al. \cite{optimal_transport_missing} addressed missing data imputation with Optimal Transport (OT). Their model computed the most efficient transformation between observed and imputed data distributions. Instead of minimizing pointwise errors, they aligned global data structures, which led to superior results in high-dimensional datasets with complex dependencies.

You et al. \cite{you2020handling} introduced graph-based models that treat the imputation problem as a network prediction task. By representing rows and features as nodes in a bipartite graph, these models propagate information across the dataset, capturing relational structure that simpler methods ignore. Likewise, Du et al. \cite{du2023remasker} proposed transformer-based imputers, which leverage self-attention to weigh the importance of all observed values when predicting a missing entry, enabling more context-aware imputations even in heterogeneous, mixed-type datasets.

\subsection{Meta-Learning and Ensemble Approaches}

Recent advancements also emphasize the need for meta-level strategies in selecting appropriate imputation techniques. Liu et al. \cite{meta1} introduced a Provenance Meta Learning Framework that leverages extensive meta-attributes (such as missingness patterns, dataset characteristics, and classification performance metrics) to automate the selection of missing data handling methods. Their framework uses rule-based learning to recommend the most effective imputation technique, addressing the issue that no single method performs best across all scenarios.
In ecological data analysis, Ellington et al. \cite{meta2} evaluated the performance of multiple imputations for meta-regression, demonstrating that it can outperform complete case analysis in producing unbiased and precise estimates. However, they also showed that its effectiveness varies depending on whether the missing data pertains to predictors, weights, or derived variables. Their work underlines the complexity of imputation performance and the importance of simulating different missingness mechanisms, principles that our meta-imputation strategy builds upon to ensure robust generalization across contexts.

Although significant progress has been made in developing imputation methods, most rely on fixed strategies that may not generalize well across datasets or types of missingness. Few studies have investigated how different imputation methods could be combined systematically and data-driven. This work proposes to fill that gap by introducing a meta-level approach that learns to select or combine multiple imputation outputs using a supervised model trained on artificially masked data. By adapting to the behavior of various imputers, the proposed framework aims to improve reliability and performance in practical, real-world scenarios.


\begin{table*}[ht]
\centering
\caption{Comparison of Imputation Approaches from Literature}
\label{tab:related_work_comparison}
\renewcommand{\arraystretch}{1.3}
\begin{tabular}{|m{3cm}|m{3.1cm}|m{4cm}|m{2.5cm}|m{2.5cm}|}
\hline
\textbf{Approach} & \textbf{Representative Methods} & \textbf{Key Idea} & \textbf{Missingness Types Addressed} & \textbf{Hypothetical Notes} \\
\hline
\multirow{3}{*}{\textbf{Traditional Statistical}} 
& Mean & \multirow{3}{4cm}{Fills missing values using simple summary statistics like average or most frequent values.} & \multirow{3}{2.5cm}{MCAR only} & \multirow{3}{2.5cm}{ Schafer \& Graham \cite{zhou2024review}, Little \cite{article}} \\
& Median & & & \\
& Mode & & & \\
\hline
\multirow{4}{*}{\textbf{Machine Learning}} 
& KNN & \multirow{4}{4cm}{Exploits relationships between features to estimate missing data values.} & \multirow{4}{2.5cm}{MCAR, partially MAR} & \multirow{4}{2.5cm}{Murti et al. \cite{KNN}, Lee et al. \cite{matrix}, Y.-Y. Song \& Y. Lu \cite{decision_tree}, Feng et al. \cite{svm_missing}} \\
& Matrix Factorization & & & \\
& Decision Trees & & & \\
& SVM & & & \\
\hline
\multirow{3}{*}{\textbf{Deep Learning}} 
& Autoencoders & \multirow{3}{4cm}{Uses neural representations and adversarial methods to impute data realistically.} & \multirow{3}{2.5cm}{MCAR, MAR (limited MNAR)} & \multirow{3}{2.5cm}{Collier et al. \cite{vae_missing}, Yoon et al. \cite{gain}, Du et al. \cite{du2023remasker} } \\
& GAIN & & & \\
& Transformers & & & \\
\hline
\multirow{2}{*}{\textbf{Meta-Learning}} 
& Provenance Meta Learning & \multirow{2}{4cm}{Combines multiple imputers based on dataset characteristics to optimize performance.} & \multirow{2}{2.5cm}{MCAR, partially MAR; potential for MNAR} & \multirow{2}{2.5cm}{Liu et al. \cite{meta1}, Ellington et al. \cite{meta2}} \\
& Meta-Regression & & & \\[0.6ex]
\hline
\end{tabular}
\end{table*}

To provide a concise overview of the methods discussed, Table~\ref{tab:related_work_comparison} summarizes the main classes of imputation approaches, representative techniques, and the types of missingness they are typically designed to handle.

\section{Meta-Imputation Balanced (MIB)}

We introduce MIB, a supervised ensemble framework that improves missing value estimation by learning how to combine the strengths of multiple imputation techniques. Instead of relying on a single method, our framework uses a meta-model to integrate the outputs of several base imputers. This approach yields more accurate, stable, adaptive imputations across various data domains.

\subsection{Overview}

MIB operates in two stages: base imputation and meta-model training. In the first stage, we independently apply a set of predefined imputation methods, namely Mean \cite{zhou2024review}, Median \cite{zhou2024review}, Mode \cite{zhou2024review}, KNN \cite{KNN}, Matrix Factorization \cite{matrix}, Autoencoders \cite{autoencoders_missing}, and GAIN \cite{gain} to the dataset. Each method generates a complete version of the dataset by filling in missing values according to its logic. These imputed results serve as candidate inputs to the second stage, where a regression-based meta-model learns to predict the true value of each masked entry based on the outputs of all base methods. The meta-model treats the set of imputed values as a feature vector and uses the ground truth (available during training) to learn an optimal combination strategy.

\subsection{Training Phase}

To train the MIB, we randomly masked 10\% of the entries under the Missing Completely at Random (MCAR) assumption, implemented by independently masking entries using values drawn from a uniform distribution and a fixed random seed for reproducibility. Each missing entry became a supervised training example, with the outputs from the base imputers serving as input features and the original (pre-masked) value as the target. We used a Linear Regression meta-model with Mean Squared Error as the optimization objective.

Let \( \mathcal{D} \in \mathbb{R}^{n \times d} \) be the complete data matrix and \( \mathcal{M} \in \{0, 1\}^{n \times d} \) be the binary mask indicating missing entries, where \( \mathcal{M}_{ij} = 1 \) if the value at position \((i, j)\) is missing. Here, \( i \in \{1, \dots, n\} \) indexes the rows (instances) and \( j \in \{1, \dots, d\} \) indexes the columns (features), so that each pair \((i, j)\) uniquely identifies an entry in the data matrix. Let \( \{ \hat{\mathcal{D}}^{(k)} \}_{k=1}^K \) denote the set of \( K \) imputed matrices produced by the base imputers.

Let \( N \) denote the total number of missing entries in the dataset, and index them with \( z = 1, \dots, N \). For each missing entry \( z \) corresponding to some position \( (i, j) \), we construct an input feature vector:
\[
\mathbf{x}_{z} = \left[ \hat{\mathcal{D}}_{ij}^{(1)}, \hat{\mathcal{D}}_{ij}^{(2)}, \ldots, \hat{\mathcal{D}}_{ij}^{(K)}; \mathbf{f}_j \right] \in \mathbb{R}^{K + d},
\]

\noindent and assign the corresponding target value:
\[
y_{z} = \mathcal{D}_{ij}.
\]

Here, \( \mathbf{f}_j \) represents the feature-level metadata associated with column \( j \), such as statistical properties or data type encodings. This allows the meta-model to incorporate information about the column being imputed in addition to the base imputers' predictions.

The meta-model \( f \) is trained by minimizing the empirical squared loss over all missing positions. In our implementation, we used a Linear Regression model as the meta-model, chosen for its simplicity, interpretability, and ability to learn optimal weights for combining the base imputers' outputs without introducing additional non-linear complexity.

\subsection{Inference Phase}

At test time, for each new missing value \( (i, j) \), we again collect imputed values from each base method to form:
\[
\mathbf{x}_{z}^{\text{test}} = \left[ \hat{\mathcal{D}}_{ij}^{(1)}, \hat{\mathcal{D}}_{ij}^{(2)}, \ldots, \hat{\mathcal{D}}_{ij}^{(K)}; \mathbf{f}_j \right],
\]

\noindent and compute the final imputed values as:
\[
\hat{y} = f\left( \mathbf{x}^{\text{test}} \right).
\]

\noindent where $\hat{y} \in \mathbb{R}^N$ denotes the vector of $N$ imputed values.

This formulation enables the MIB to learn a data-driven, context-aware strategy for combining imputation methods, providing more accurate and adaptive estimates than fixed or rule-based approaches. During inference, the same set of base imputers generates their candidate values for each missing entry. These values are passed to the trained Linear Regression meta-model, which outputs the final imputed value. This simple but effective combination allows the model to weigh each imputer differently depending on the context of the missing value and feature (variable).


\section{Evaluation}

We evaluate MIB through a series of experiments designed to measure the performance of imputed values and their impact on downstream predictive performance. Our evaluation uses controlled missingness and standard benchmark datasets to ensure consistency and reproducibility.

\subsection{Experimental Setup and Baselines}

We evaluated our method using 5-fold cross-validation to obtain robust and unbiased performance estimates. In each fold, four subsets were used for training and one for testing, with shuffling applied before splitting. Artificial gaps were introduced by randomly masking a specified fraction of entries (10\% in our study) in both training and testing data under the MCAR assumption. We selected 10\% missingness as it is a common benchmark in imputation studies, providing sufficient artificial gaps to test robustness without overly distorting the datasets.

We implement all experiments in Python using scikit-learn \cite{scikit-learn}, XGBoost \cite{xgboost}, Linear Regression \cite{Linear_regression}, and custom modules that support each imputation strategy.
We compare MIB with a set of widely used baseline methods that reflect a range of imputation philosophies. These include simple statistical techniques (mean, median, and mode), proximity-based algorithms (such as k-nearest neighbors), predictive model-driven approaches (matrix factorization), and deep learning models (autoencoders and GAIN). Each baseline runs independently on the same data and under the same masking and preprocessing conditions to allow for direct comparison.

\subsection{Evaluation Metrics and Expectations}

Metrics are averaged over the five folds to account for variability across different train/test partitions. We assess each imputation method using two complementary perspectives:
\begin{itemize}
\item Direct Evaluation: This evaluation computes the MAE and RMSE on the artificially masked entries by comparing the imputed values to the original ground truth. These metrics quantify the raw performance of each imputation method.
\item Indirect Evaluation: Supervised machine learning models (Random Forest \cite{Random_forest}, XGBoost\cite{xgboost}, and Linear Regression \cite{Linear_regression}) are trained on the completed datasets to predict a designated target variable. We use the RMSE on a test set for regression tasks to evaluate how well the imputed data preserves the performance of the original model trained on the full dataset.
\end{itemize}

\subsection{Datasets}
\label{sec:datasets}

We used three publicly available tabular datasets to evaluate the MIB framework:

\begin{itemize}
  \item Diabetes Health Indicators (Kaggle)\cite{diabetes_kaggle}: A large-scale dataset with over 250,000 examples and more than 20 features, including demographics, behavioral risk factors, and general health indicators, commonly used in public health prediction tasks.
  \item Heart Disease (Kaggle) \cite{heart_kaggle}: A mid-size dataset related to cardiovascular health and diagnosis.  Its moderate size (1025 examples) makes it well‑suited to reveal over‑smoothing by statistical imputers or overfitting by complex models.
  \item Gallstone (UCI) \cite{gallstone_uci}: A small clinical dataset with 320 instances, containing demographic and medical features associated with gallstone disease. Its limited size and mixed variable types present a more challenging imputation environment, particularly for deep‑learning models.
\end{itemize}

These datasets differ in size, noise levels, and feature types, making them suitable for evaluating both simple and advanced imputation methods.

\section{Results}


We evaluated MIB on the three datasets described in Section~\ref{sec:datasets}, comparing its performance with a range of baseline imputation methods.

All baseline imputers were evaluated under identical conditions. We measured:

\begin{itemize}
  \item Direct performance using MAE and Masked RMSE between imputed and true values; and
  \item Indirect performance via Prediction RMSE obtained by training Random Forest, XGBoost, and Linear Regression models on the imputed datasets.
\end{itemize}

\subsection{Diabetes Health Indicators Dataset}

As shown in Table~\ref{tab:diabetes}, MIB achieved a competitive RMSE (0.975), second only to XGBoost (0.938), while maintaining a lower MAE (0.697) than most single-model imputers except Median (0.633), KNN (0.642), and XGBoost (0.651). This indicates only a minor trade-off for stability across models. Notably, deep learning approaches (GAIN, Autoencoder) did not outperform simpler methods, both showing RMSE values above 1. For downstream prediction, MIB achieved the best RMSE for Random Forest (0.882) and remained close to the top for XGBoost and Linear Regression.

\begin{table*}[h!]
\caption{Evaluation of imputation methods on the \textit{Diabetes Health Indicators} dataset. Masked MAE and RMSE measure direct imputation performance; Prediction RMSE reflects downstream model performance. RF = Random Forest, XGB = XGBoost, LR = Linear Regression.}
\begin{center}
\label{tab:diabetes}
\begin{tabular}{l|cc|ccc}
\hline
 & \multicolumn{2}{c|}{\textbf{Direct}} & \multicolumn{3}{c}{\textbf{Indirect}} \\
\hline
Imputation Method & 
\makecell{Masked\\MAE} & 
\makecell{Masked\\RMSE} & 
\makecell{Prediction\\RMSE (RF)} & 
\makecell{Prediction\\RMSE (XGB)} & 
\makecell{Prediction\\RMSE (LR)} \\
\hline
Mean & 0.770 & 1.016 & 0.886 & 0.963 & 0.856 \\
Median & \textbf{0.633} & 1.168 & 0.902 & 0.997 & 0.868 \\
Mode & 0.667 & 1.219 & 0.902 & 0.986 & 0.868 \\
KNN & 0.642 & 0.990 & 0.890 & 0.967 & 0.856 \\
XGBoost & 0.651 & \textbf{0.938} & 0.869 & \textbf{0.934} & \textbf{0.846} \\
Matrix Factorization & 0.857 & 1.395 & 0.958 & 1.052 & 0.924 \\
Autoencoder & 0.823 & 1.130 & 0.896 & 0.969 & 0.862 \\
GAIN & 0.771 & 1.016 & 0.888 & 0.960 & 0.856 \\
\textbf{MIB (Ours)} & 0.697 & 0.975 & \textbf{0.882} & 0.953 & 0.852 \\
\hline
\end{tabular}
\end{center}
\end{table*}

\subsection{Heart Disease Dataset}

Table~\ref{tab:heart} shows that, in this medium-sized dataset, MIB delivered the lowest MAE (0.499) and the lowest RMSE (0.702), outperforming all baselines on both direct metrics. Traditional imputers (Mean, Median, Mode) performed worse (RMSE $\geq$ 0.979), and Matrix Factorization and deep learning methods struggled even more. MIB also achieved the second-best Prediction RMSE for Random Forest (0.786), slightly above the best score from XGBoost (0.782), and a competitive result with other models, confirming its stability and generalizability across model types.

\begin{table*}[h!]
\caption{Evaluation of imputation methods on the \textit{Heart Disease} dataset. Masked MAE and RMSE measure direct imputation performance; Prediction RMSE reflects downstream model performance. RF = Random Forest, XGB = XGBoost, LR = Linear Regression.}
\begin{center}
\label{tab:heart}
\begin{tabular}{l|cc|ccc}
\hline
 & \multicolumn{2}{c|}{\textbf{Direct}} & \multicolumn{3}{c}{\textbf{Indirect}} \\
\hline
Imputation Method & 
\makecell{Masked\\MAE} & 
\makecell{Masked\\RMSE} & 
\makecell{Prediction\\RMSE (RF)} & 
\makecell{Prediction\\RMSE (XGB)} & 
\makecell{Prediction\\RMSE (LR)} \\
\hline
Mean & 0.863 & 0.979 & 0.834 & 0.842 & 0.908 \\
Median & 0.772 & 1.083 & 0.851 & 0.880 & 0.915 \\
Mode & 0.760 & 1.149 & 0.830 & 0.836 & 0.908 \\
KNN & 0.551 & 0.723 & 0.804 & 0.869 & 0.907 \\
XGBoost & 0.539 & 0.712 & \textbf{0.782} & \textbf{0.768} & \textbf{0.893} \\
Matrix Factorization & 0.975 & 1.373 & 0.913 & 0.947 & 0.987 \\
Autoencoder &1.055 & 1.326 & 0.922 & 0.963 & 1.004 \\
GAIN & 0.861 & 0.976 & 0.834 & 0.827 & 0.907 \\
\textbf{MIB (Ours)} & \textbf{0.499} & \textbf{0.702} & 0.786 & 0.872 & 0.907 \\
\hline
\end{tabular}
\end{center}
\end{table*}

\subsection{Gallstone Dataset}

As illustrated in Table~\ref{tab:gallstone}, MIB obtained an RMSE of 1.004, slightly higher than the best (XGBoost at 0.804 and KNN at 0.833), and an MAE (0.711) close to KNN (0.500) and XGBoost (0.461). Deep learning models again underperformed, with Autoencoder and GAIN yielding RMSEs above 1. Importantly, MIB achieved the lowest Prediction RMSE (2.077) under Linear Regression, outperforming all baselines on this metric, and maintained stable performance across Random Forest and XGBoost.

\begin{table*}[h!]
\caption{Evaluation of imputation methods on the \textit{Gallstone} dataset. Masked MAE and RMSE measure direct imputation performance; Prediction RMSE reflects downstream model performance. RF = Random Forest, XGB = XGBoost, LR = Linear Regression.}
\begin{center}
\label{tab:gallstone}
\begin{tabular}{l|cc|ccc}
\hline
 & \multicolumn{2}{c|}{\textbf{Direct}} & \multicolumn{3}{c}{\textbf{Indirect}} \\
\hline
Imputation Method & 
\makecell{Masked\\MAE} & 
\makecell{Masked\\RMSE} & 
\makecell{Prediction\\RMSE (RF)} & 
\makecell{Prediction\\RMSE (XGB)} & 
\makecell{Prediction\\RMSE (LR)} \\
\hline
Mean & 0.708 & 1.001 & 0.852 & 0.950 & 2.927 \\
Median & 0.673 & 1.027 & 0.835 & 0.923 & 2.100 \\
Mode & 0.814 & 1.203 & 0.852 & 1.032 & 2.104 \\
KNN & 0.500 & 0.833 & 0.827 & 0.890 & 1.700 \\
XGBoost & \textbf{0.461} & \textbf{0.804} & \textbf{0.813} & \textbf{0.813} & 11.840 \\
Matrix Factorization & 0.981 & 1.336 & 0.845 & 0.966 & 4.202 \\
Autoencoder & 1.125 & 1.578 & 0.862 & 0.968 & 2.002 \\
GAIN & 0.729 & 1.024 & 0.875 & 0.982 & 3.356 \\
\textbf{MIB (Ours)} & 0.711 & 1.004 & 0.828 & 0.942 & \textbf{2.077} \\
\hline
\end{tabular}
\end{center}
\end{table*}

\vspace{1em}
 In summary, while individual methods occasionally produced the lowest MAE or RMSE on a given dataset, MIB consistently achieved top-tier performance and, in several cases, the best results across direct or indirect metrics. Even when not producing the absolute lowest RMSE (e.g., against XGBoost in the Diabetes and Gallstone datasets), MIB maintained competitive performance while delivering the most stable performance across datasets, metrics, and downstream models. These results validate the strength of the MIB approach in providing robust, generalizable imputations for diverse tabular datasets.


        
        

\section{Conclusion}

In this study, we addressed the persistent challenge of missing data in machine learning by proposing a MIB framework. Unlike traditional single-method approaches, our method leverages ensemble learning to combine multiple imputation strategies, including both statistical and machine learning-based techniques. Through a supervised meta-model trained on synthetically masked data, our system learns to predict the most suitable imputed values by capturing the strengths of individual base imputers.

Our evaluation across three benchmark datasets, Diabetes Health Indicators, Heart Disease, and Gallstone, showed that the proposed framework consistently delivered the best or near-best RMSE while maintaining competitive MAE across all settings. While certain single-method imputers (e.g., Median or XGBoost) occasionally achieved the lowest MAE, MIB offered the most stable RMSE performance and robust downstream predictive performance across Random Forest, XGBoost, and Linear Regression models. These findings highlight that even a simple, interpretable meta‑model can successfully integrate multiple imputers into a unified system, achieving generalizable performance without the computational burden of deep or black-box approaches.

While it is easy to pick the best single method after testing, it would be much more difficult to do this before even training the model. Typically, this would require tuning an additional hyperparameter (imputation method). It is much more straightforward to rely on the parameter-less MIB method, which consistently performed among the best.

Nevertheless, several limitations remain. Our experiments focused on the MCAR assumption, and the framework's adaptability to more complex missingness types, such as MAR and MNAR, remains an area for future exploration. In future work, we also intend to investigate the impact of varying higher levels of missingness, beyond the 10\% considered in this study. It is also important to note that the current framework is limited to structured, numerical data; extending it to support unstructured and categorical data would require substantial methodological adjustments and remains an open avenue for future research.

Ultimately, this work highlights the potential of ensemble-based, meta-level strategies for handling missing data. Rather than relying on a single imputation method, our results demonstrate that a flexible, modular approach can adapt across datasets and missingness patterns, making it a promising building block for robust, real-world machine learning pipelines.

\section*{Acknowledgement}
The research and development presented in this paper were funded by the Research Agency of the Republic of Slovenia (ARIS) through the ARIS Young Researcher Programme (research core funding No. P2-209).
While preparing this work, the authors used Grammarly to check the correctness of grammar and improve the fluency of the writing, aiming to enhance the clarity and impact of the publication. The authors reviewed and edited the content produced with this tool/service and accept full responsibility for the final published content.


\bibliographystyle{IEEEtran}
\bibliography{references}

\end{document}